\documentclass[letterpaper, 10 pt, conference]{ieeeconf}  

\IEEEoverridecommandlockouts                              

\usepackage{amsmath,amsfonts}
\usepackage{algorithmic}
\usepackage{algorithm}
\usepackage{array}
\usepackage[caption=false,font=normalsize,labelfont=sf,textfont=sf]{subfig}
\usepackage{textcomp}
\usepackage{stfloats}
\usepackage{url}
\usepackage{verbatim}
\usepackage{graphicx}
\usepackage{xcolor}
\usepackage{colortbl}
\usepackage{amsmath}
\usepackage{amssymb}
\usepackage{ctable} 
\usepackage{multirow}
\usepackage{multicol}

\title{\LARGE \bf
DISCOVERSE: Efficient Robot Simulation in \\ Complex High-Fidelity Environments
}

\author{Yufei Jia$^{1, \dagger}$, Guangyu Wang$^{1, \dagger}$, Yuhang Dong$^{2}$, Junzhe Wu$^{1}$, Yupei Zeng$^{2}$, Haonan Lin$^{3}$, Zifan Wang$^{4}$, \\ Haizhou Ge$^{1}$, Weibin Gu$^{1}$, Kairui Ding$^{1}$, Zike Yan$^{1}$, Yunjie Cheng$^{5}$, Yue Li$^{7}$, Ziming Wang$^{6}$, Chuxuan Li$^{1}$, \\ Wei Sui$^{8}$, Lu Shi$^{1}$, Guanzhong Tian$^{2}$, Ruqi Huang$^{1, \ddagger}$, Guyue Zhou$^{1, \ddagger}$
\thanks{$^{{1}}$Tsinghua University, $^{{2}}$Zhejiang University, $^{{3}}$Huazhong University of Science and Technology, $^{{4}}$Hong Kong University of Science and Technology (Guangzhou), $^{{5}}$Xi'an Jiaotong University, $^{{6}}$Tongji University, $^{{7}}$DISCOVER Robotics, $^{{8}}$D-Robotics.}
\thanks{{$^\dagger$Yufei Jia and Guangyu Wang contributed equally to this work (email: \{jyf23, wanggy24\}@mails.tsinghua.edu.cn).}}
\thanks{$^\ddagger$Corresponding authors: Ruqi Huang; Guyue Zhou (email: ruqihuang@sz.tsinghua.edu.cn; zhouguyue@air.tsinghua.edu.cn).}
}

\begin{document}

\maketitle
\thispagestyle{empty}
\pagestyle{empty}

\begin{abstract}
We present \textsc{Discoverse}, 
the first unified, modular, open-source 3DGS-based simulation framework for Real2Sim2Real robot learning. 
It features a holistic Real2Sim pipeline that synthesizes hyper-realistic geometry and appearance of complex real-world scenarios, paving the way for analyzing and bridging the Sim2Real gap. Powered by Gaussian Splatting and MuJoCo, \textsc{Discoverse} enables massively parallel simulation of multiple sensor modalities and accurate physics, with inclusive supports for existing 3D assets, robot models, and ROS plugins, empowering large-scale robot learning and complex robotic benchmarks. Through extensive experiments on imitation learning, \textsc{Discoverse} demonstrates state-of-the-art zero-shot Sim2Real transfer performance compared to existing simulators. 
For code and demos: https://air-discoverse.github.io/.
\end{abstract}

\section{Introduction}
End-to-end learning has emerged as a scalable, efficient, and cost-effective solution for robotics,
enabling direct policy learning from raw sensor data. This paradigm underscores the crucial need for fast and robust simulators, a research area that has witnessed rapid advancements in recent years~\cite{todorov2012mujoco, szot2021habitat, mittal2023orbit, xiang2020sapien, gan2020threedworld, mu2024robotwin, gu2023maniskill2, ehsani2021manipulathor, li2021igibson}. However, a critical challenge in end-to-end learning remains largely unsolved, i.e., the dramatic performance degradation when transferring from simulation to the real world, a.k.a. the Sim2Real gap~\cite{hofer2021sim2real}, which fundamentally originates from the visual discrepancies between simulation and reality~\cite{gervet2023navigating}. In simulation, the appearance is typically rendered from artificial assets with handcrafted textures and simplified lighting, failing to faithfully characterize the complexity of the real world. 

To circumvent this issue, some Real2Sim approaches~\cite{chang2017matterport3d, straub2019replica, savva2019habitat, xia2020interactive} leverage 3D reconstruction techniques to build virtual replicas of the real world, with the aim to inherit the rich appearance, structures, and semantics from reality. However, they are mainly designed for navigation-oriented tasks and primarily lack photorealism. The reason is that traditional multi-view stereo (MVS) and RGB-D fusion approaches are vulnerable to non-Lambertian reflectance and thin geometry, inevitably leading to large amounts of collapsed surface and thus severely deteriorated visual quality. On the other hand, state-of-the-art simulator Omniverse Issac Lab~\cite{mittal2023orbit} enables high-quality Physically-Based Rendering (PBR) in real-time through GPU acceleration. Despite the notable progress, they need onerous configurations and lack supports for Real2Sim assets, especially for those with large amounts of primitives. Very recently, there are several early attempts~\cite{li2024robogsim, lou2024robo, Qureshi2024SplatSimZS} that utilize advanced neural representations -- e.g., 3D gaussian splatting (3DGS)~\cite{kerbl20233d} -- to build Real2Sim replicas as radiance fields. However, they fail to recover precise geometry and relightable appearance, exhibit poor view extrapolation capabilities, and lack robustness in complex real-world scenarios. For example, they struggle to handle in-the-wild scenes with intricate geometry, textures, and illumination, large-scale scenes with unstructured and sparse imagery, and also textureless or non-Lambertian surfaces. Therefore, they are unsuitable as general-purpose robotic simulators for diverse real-world applications.

\begin{figure*}[htbp]
    \centering
    \newcommand{\colw}{0.19}
    \newcommand{\figw}{1} 
    \includegraphics[width=\figw\textwidth,trim={0cm 0cm 0cm 0cm},clip]{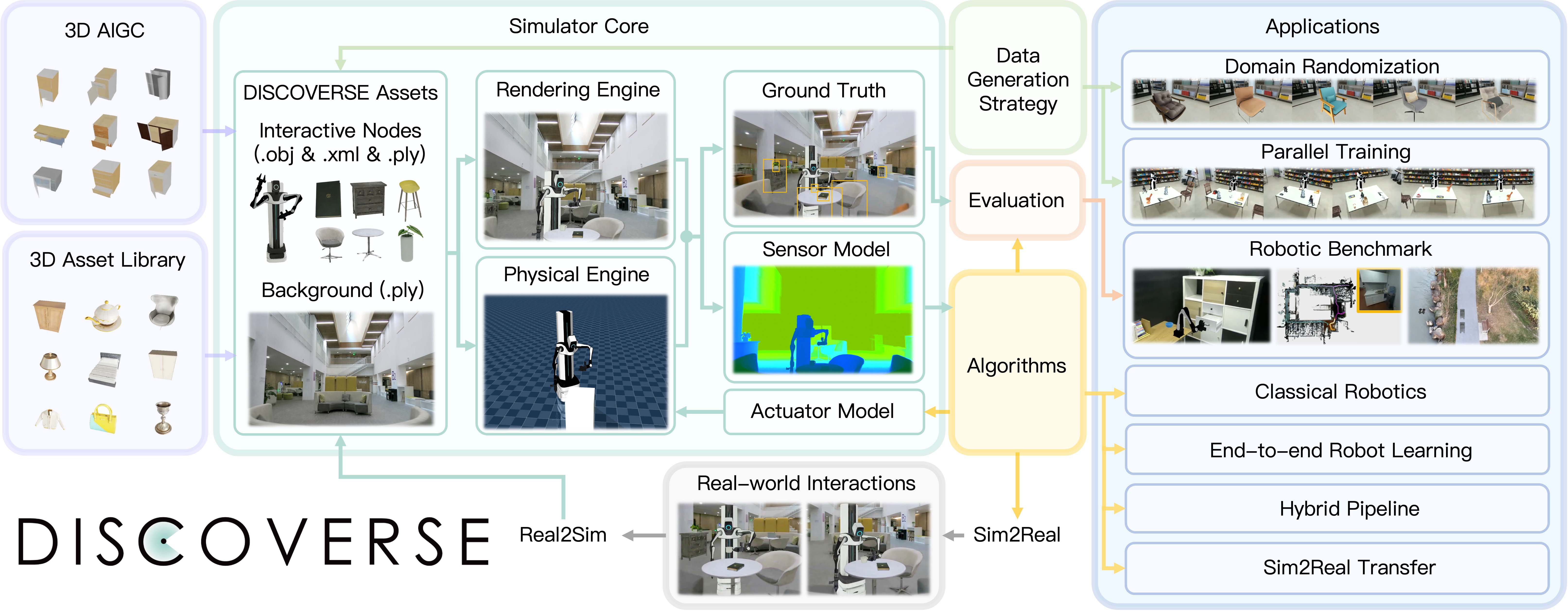}
    \hfill
\vspace{-0.6cm}
\caption{\textsc{Discoverse} system overview. \textsc{Discoverse} unifies real-world captures, 3D AIGC, and any existing 3D assets in formats of 3DGS (.ply), mesh (.obj/.stl), and MJCF physical models (.xml), enabling their use as interactive scene nodes (objects and robots) or the background node. 
We leverage Gaussian splatting as our rendering engine to generate hyper-realistic radiance field rendering of multiple sensor modalities and use MuJoCo as the physical engine to ensure accurate physics.
Benefiting from the efficiency and fidelity, \textsc{Discoverse} enables user-definable data generation strategy, evaluation metrics, and algorithms for robotics and embodied AI, empowering a variety of applications, e.g., parallel training, complex robotic benchmarks, etc.}
\label{fig:teaser}
\end{figure*}

In light of these observations, we introduce {\textsc{Discoverse}}, the first unified, modular, open-source 3DGS-based simulation framework with a collection of novel features to facilitate end-to-end robotic solutions: 
\begin{enumerate}
\item{High-fidelity, hierarchical Real2Sim generation for both background node and interactive scene nodes in various complex real-world scenarios, leveraging advanced laser-scanning, generative models, physically-based relighting, and Mesh-Gaussian transfer.
}
\item{Efficient simulation and user-friendly configuration. By seamlessly integrating 3DGS rendering engine, MuJoCo physical engine, and ROS2 robotic interface, we provide an easy-to-use, massively parallel implementation for rapid deployment and flexible extension. We also propose an automated state generation approach to facilitate demonstration collection. The overall throughput achieves 650 FPS for 5 cameras rendering RGB-D frames, which is $\sim$3$\times$ faster than Issac Lab (ORBIT)~\cite{mittal2023orbit}.}

\item{Compatibilities with existing
3D assets and inclusive supports for robot models (robotic arm, mobile manipulator, quadrocopter, etc.), sensor modalities (RGB, depth, LiDAR, tactile sensors), ROS plugins, and various randomizations (e.g., generative-based).
}
\end{enumerate}

A high-level comparison with existing simulators is listed in Table~\ref{tab:sim_compare}. Unlike prior works, \textsc{Discoverse} offers a more comprehensive Real2Sim solution that generates high-quality replicas of diverse real-world scenarios with precise geometry and harmonious appearance, 
thus yielding a more robust and versatile framework for bridging the Sim2Real gap.


To verify the effectiveness of \textsc{Discoverse}, we conduct extensive experiments on imitation learning (IL) using both ACT~\cite{zhao2023learning} and Diffusion Policy (DP)~\cite{chi2023diffusion} across three real-world manipulation tasks. We compare \textsc{Discoverse} against three state-of-the-art simulators -- MuJoCo~\cite{todorov2012mujoco}, RoboTwin~\cite{mu2024robotwin}, and SplatSim~\cite{Qureshi2024SplatSimZS} -- by fairly deploying ACT and DP on each of them. Our results demonstrate that \textsc{Discoverse} significantly outperforms existing simulators in zero-shot Sim2Real transfer. In particular, we identify the following findings:
\begin{enumerate}
\item Using ACT, \textsc{Discoverse} increases the average success rate relative to the second best simulator (SplatSim) by $\sim$11\% without data augmentation, and by 18.5\% after data augmentation.
\item Similar results are demonstrated with DP, where \textsc{Discoverse} outperforms SplatSim by $\sim$11\% without data augmentation, and by 11.4\% after augmentation.
\item Image-based augmentation further mitigates domain shifts and improves the average success rate of \textsc{Discoverse} by 31.5\% with ACT and by 29.3\% with DP. \item \textsc{Discoverse} enables $\sim$100$\times$ more efficient data collection compared to real-world demonstrations.
\end{enumerate}
These results underscore the great potential of \textsc{Discoverse} in steering Sim2Real towards Real2Real. We believe \textsc{Discoverse} lays solid foundation for comprehensive Sim2Real robotic benchmarks, including manipulation, navigation, multi-agent collaboration, etc., to stimulate further research and practical applications. 

\begin{table*}[htbp]
\centering
\setlength{\tabcolsep}{2.0mm}{
\caption{Comparison between \textsc{Discoverse} and other simulators supporting end-to-end robot learning.}
\begin{tabular}{c|ccc|cccccc}
\specialrule{.1em}{.05em}{.05em}
Simulators & \multicolumn{1}{c}{\begin{tabular}[c]{@{}c@{}}3D \\ Representation\end{tabular}} & \multicolumn{1}{c}{\begin{tabular}[c]{@{}c@{}}Physics \\ Engine\end{tabular}} & Renderer & \multicolumn{1}{c}{\begin{tabular}[c]{@{}c@{}}Scene-level \\ Real2Sim\end{tabular}} & \multicolumn{1}{c}{\begin{tabular}[c]{@{}c@{}}Scene-level \\ Fidelity\end{tabular}} & \multicolumn{1}{c}{\begin{tabular}[c]{@{}c@{}}$^{\dagger}$Scene \\ Complexity\end{tabular}} & \multicolumn{1}{c}{\begin{tabular}[c]{@{}c@{}}Object-level \\ Real2Sim\end{tabular}} & \multicolumn{1}{c}{\begin{tabular}[c]{@{}c@{}}Object-level \\ Fidelity\end{tabular}} \\ 
\hline\hline
Matterport3D~\cite{chang2017matterport3d} & Mesh & {None} & MeshRender & {$\checkmark$} & {$\star$} & H & {$\times$} & {$\star$} \\
SAPIEN~\cite{xiang2020sapien} & Mesh & PhysX4 & OpenGL & {$\times$} & {$\star$} & H & {$\times$} & {$\star$} \\
ThreeDWorld~\cite{gan2020threedworld} & Mesh & PhysX4/FleX & Unity3D & {$\times$} & {$\star$} & THW & {$\times$} & {$\star$} \\
ManipulatorThor~\cite{ehsani2021manipulathor} & Mesh & PhysX4 & Unity & {$\times$} & {$\star$} & H & {$\times$} & {$\star$} \\
iGibson~\cite{xia2020interactive} & Mesh & Bullet & OpenGL & {$\checkmark$} & {$\star$} & H & {$\times$} & {$\star$} \\
Habitat2.0~\cite{szot2021habitat} & Mesh & Bullet & Magnum & {$\times$} & {$\star$} & H & {$\times$} & {$\star$} \\
ManiSkill2~\cite{gu2023maniskill2} & Mesh & PhysX4/Warp & OpenGL & {$\times$} & {$\star$} & TH & {$\times$} & {$\star$} \\
ORBIT~\cite{mittal2023orbit} (Issac Lab) & Mesh & PhysX5.1 & Omni.RTX & {$\times$} & {$\star$} & THW & {$\times$} & {$\star$} \\ 
RoboTwin~\cite{mu2024robotwin} & Mesh & PhysX4 & OpenGL & {$\times$} & {$\star$} & T & {$\checkmark$} & {$\star\star$} \\
SplatSim~\cite{Qureshi2024SplatSimZS} & 3DGS & Bullet & Splatting & {$\checkmark$} & {$\star\star$} & T & {$\checkmark$} & {$\star\star$} \\
\specialrule{.1em}{.05em}{.05em}
\textbf{{\textsc{Discoverse}} (Ours)} & 3DGS & MuJoCo & Splatting & {$\checkmark$} & {$\star\star\star$} & THW & {$\checkmark$} & {$\star\star\star$} \\ 
\specialrule{.1em}{.05em}{.05em}
\end{tabular}
}



\begin{minipage}{20cm}
\vspace{0.1cm}
\scriptsize 
\hspace{0.1cm}
${\dagger}$ For Scene Complexity, `T' stands for table-top, `H' stands for house-scale, `W' stands for in-the-wild large-scale scenes.
\end{minipage}

\label{tab:sim_compare}
\end{table*}


\section{Related Works} 

\subsection{Simulation Environments for Robotics}
Simulators~\cite{todorov2012mujoco, szot2021habitat, mittal2023orbit, xiang2020sapien, gan2020threedworld, mu2024robotwin, gu2023maniskill2, ehsani2021manipulathor, li2021igibson} are crucial in scaling up robot learning by offering efficiency and safety. The underlying physical engines~\cite{todorov2012mujoco, coumans2016pybullet} enable high-throughput simulation of accurate contact, collision, and deformation dynamics, empowering a variety of complex robotic tasks~\cite{gervet2023navigating, zhao2023learning, li2024activesplat}. As another core component, the renderer, which synthesizes visual inputs for robots, is commonly based on game engines that are widely deployed in prior frameworks~\cite{mittal2023orbit, xiang2020sapien, gan2020threedworld, mu2024robotwin, gu2023maniskill2, ehsani2021manipulathor}. While game engines are directly compatible with artificial game assets and allow for manual customization of the environments, the process of asset creation can be time-consuming, and the renderings fail to fully characterize real-world richness, thus limiting the scalability, diversity, and fidelity of simulation.

\subsection{Real2Sim2Real Robot Learning}
To alleviate domain shifts, some methods~\cite{chang2017matterport3d, savva2019habitat, li2021igibson, xia2020interactive} propose to leverage scanned 3D mesh of real-world scenes in simulation. RoboTwin~\cite{mu2024robotwin} uses 3D generation~\cite{zhang2024clay} to enable superior object-level Real2Sim quality.
Some recent works~\cite{li2024robogsim, lou2024robo, Qureshi2024SplatSimZS} further adopt 3DGS~\cite{kerbl20233d} representation to enable comprehensive 3D reconstruction. 
Despite these advancements, existing 3DGS-based solutions still face challenges in generalizing to complex real-world scenarios, since high-quality results can only be obtained with ultra-dense multi-view captures of targets featuring rich textures and diffuse reflectance. Without these conditions, the reconstruction tends to suffer from artifacts such as excessive floaters or blurriness. Besides, these simulators reconstruct interactive targets on scene and lack supports for existing mesh assets and appearance relighting, making them inefficient and less practical for general-purpose simulation. 
\section{System Architecture}
\label{sec:system}
Our goal is to integrate an advanced neural renderer, a state-of-the-art physical engine, and a user-friendly robotic interface into a unified, modular framework, to support various end-to-end robotic perception and interaction tasks (Fig.~\ref{fig:teaser}). The simulator is developed with highly optimized implementations and offers Python API to enable rapid deployments and flexible user-driven extensions. 

In the following, we introduce the key components of our simulation framework, focusing on two main perspectives: the \textsc{Discoverse} engine and \textsc{Discoverse} asset. We then report our simulation throughput accordingly.

\begin{figure}[htbp]
    \centering
    \newcommand{\colw}{0.19}
    \newcommand{\figw}{0.48} 
    \includegraphics[width=\figw\textwidth,trim={0cm 0cm 0cm 0cm},clip]{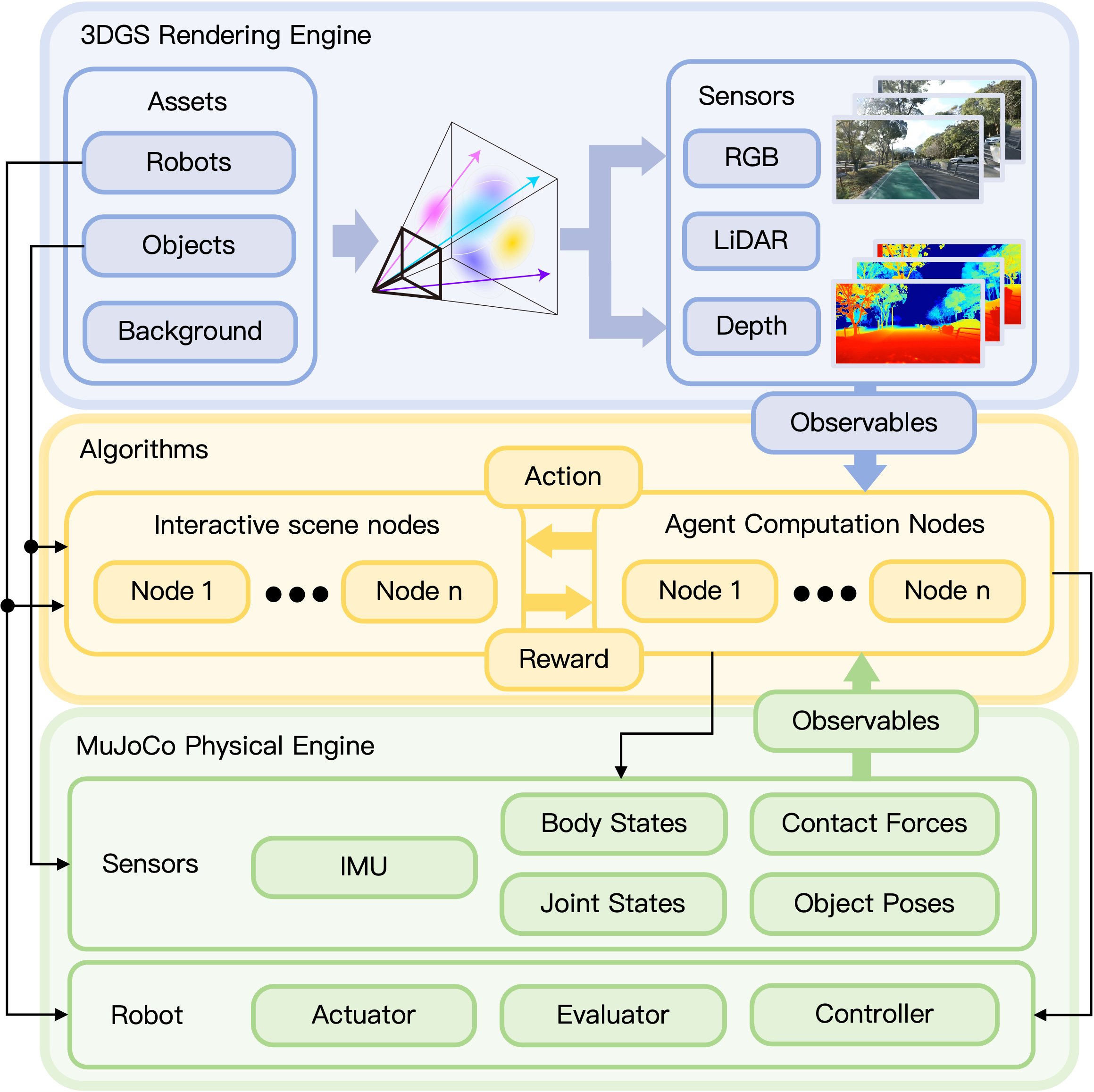}
    \hfill
\vspace{-0.5cm}
\caption{\textsc{Discoverse} operation flow. We utilize fast tile-based splatting for high-fidelity neural rendering and integrate MuJoCo~\cite{todorov2012mujoco} physical simulator for various robotic utilities. 
}
\vspace{-0.3cm}
\label{fig:flow}
\end{figure}
\subsection{Engine}
We adopt the tile-based renderer in 3D Gaussian Splatting~\cite{kerbl20233d} to simulate high-fidelity visuals and use the open-source MuJoCo~\cite{todorov2012mujoco} physical engine for accurate robot-object interactions. Our system also includes easy-to-use ROS2 (Robot Operating System 2)~\cite{macenski2022robot} support, to facilitate seamless integration and enhance robotic research workflows. The overall operation flow of \textsc{Discoverse} is depicted in Fig.~\ref{fig:flow}.

{\bf{Renderer.}} To strike a good balance between efficiency and fidelity, we use the native 3DGS renderer~\cite{kerbl20233d} in \textsc{Discoverse}, which features a fast tile-based Gaussian rasterizer with efficient sorting, splatting, spherical harmonics evaluation, and alpha-blending. All of the rendering operations are implemented as highly-optimized custom CUDA kernels to maximize GPU parallelization. 

{\bf{Physical Simulator.}} We choose MuJoCo~\cite{todorov2012mujoco} for rigid body simulation due to its efficiency and accuracy in handling complex physical interactions, including contacts, friction, and soft constraints. Leveraging MuJoCo’s capabilities, \textsc{Discoverse} supports precise force control, P-D control, and inverse dynamics, enabling accurate collision detection, contact force simulation, and the modeling of articulated robot arms or grippers with various joint types under both kinematic and dynamic control.

{\bf{ROS2 Interface.}} ROS2 (Robot Operating System 2)~\cite{macenski2022robot} is a modular and flexible framework for building robotic applications, offering advanced real-time capabilities, scalability, security, and interoperability. The ROS2 interface in \textsc{Discoverse} offers a set of APIs for interacting with robots using physics. Specifically, we apply encoder torques on joints for low-level control and provide joint space and Cartesian coordinate space APIs for higher-level control, thus bridging the gap between simulation and real-world robotic applications.

\subsection{Asset}
\label{system:asset}
High-quality digital assets are central to simulation. To seamlessly integrate the Gaussian splatting renderer with the mesh-based MuJoCo physical engine while ensuring compatibility with existing asset libraries, we introduce a comprehensive Real2Sim pipeline (detailed in Sec.~\ref{sec:real2sim}). 
With this functionality, we can easily incorporate real-world captures, 3D AIGC, and existing 3D asset libraries into \textsc{Discoverse}. Additionally, we provide full support for a variety of robot models tailored to different downstream tasks.

{\bf{Real-world Captures.}} Our simulator supports multi-view captures of real-world scenes and objects, which are processed using the established workflows of COLMAP~\cite{schonberger2016pixelwise} and 3DGS~\cite{kerbl20233d, Qureshi2024SplatSimZS}. We also have extended supports for laser scanners, so as to enable more robust and accurate reconstructions via the proposed Real2Sim functionality.

{\bf{AIGC and Asset Library.}} We ensure compatibilities with recent state-of-the-art 3D AIGC techniques, including textured mesh generation~\cite{xiang2024structured, zhang2024clay}, native 3DGS generation~\cite{tang2025lgm}, and the generation of articulated objects. We also have supports for public 3D datasets, e.g., ShapeNet~\cite{chang2015shapenet}, PartNet~\cite{mo2019partnet}, Objaverse~\cite{deitke2024objaverse}, etc. The universal support for existing 3D assets makes \textsc{Discoverse} a versatile robotic simulator for diverse application scenarios. 

{\bf{Robot Models.}} We offer a wide range of fully functional robotic agents with real robot platforms, including a robotic arm (AIRBOT Play), a dual-arm humanoid mobile manipulator (AIRBOT MMK2), a wheeled loco-manipulator, a quadrocopter, and other commonly used simulation agents~\cite{todorov2012mujoco, mittal2023orbit}. The diversity of embodiment types enables a comprehensive exploration of various perception and interaction capabilities in \textsc{Discoverse}.

{\bf{Asset Formatting.}} For interactive scene nodes, i.e., interactive objects and agents, we utilize a dual 3DGS-Mesh representation, where 3DGS (.ply) generates high-fidelity visuals, and the mesh counterpart (.obj/.stl), further described in the MJCF scene description language, ensures accurate physics simulation. Each interactive object or agent model in \textsc{Discoverse} is linked to a corresponding {.xml} file, which can be efficiently loaded into the simulation. We either randomize or manually adjust physical properties, such as friction, damping, and density, within appropriate ranges. To ensure accurate contact simulation, we decompose meshes into convex parts~\cite{wei2022approximate}. For robot models, Python APIs are provided to assemble the robot piece by piece using URDF. For background nodes, we construct only the native 3DGS representation for rendering, as no physical interaction is required.

\subsection{Sensor}
\textsc{Discoverse} enables high-fidelity simulation of diverse sensor modalities, including rendering-based sensors (RGB, depth, LiDAR) and physics-based sensors (contact force, body and joint states, IMU, and optical tactile sensor). Specifically, we utilize Gaussian splatting to enable direct simulation of RGB and depth by alpha-compositing the color and z-position of the intersected Gaussian primitives. We further propose a BVH-accelerated, native Gaussian ray tracing framework to enable highly efficient LiDAR simulation ($>$100 FPS).
We integrate Tacchi~\cite{chen2023tacchi} for optical tactile simulation and rely on MuJoCo for other physics-based sensors.

\subsection{Actuator}
Our actuator model integrates the framework from MuJoCo, incorporating control inputs and optional activation states to represent muscles and pneumatic cylinders with first-order dynamics. The model supports fixed or state-dependent gains, capturing force-length-velocity properties of muscles. Actuators transmit forces to the multi-joint system through direct joint actuation, tendon-driven mechanisms, or slider-crank systems that convert linear motion into angular movement.


\subsection{Throughput Report}

Due to its massive parallelization capabilities, \textsc{Discoverse} achieves a total of 650 FPS for rendering hyper-realistic RGB-D frames at 640$\times$480 resolution with 5 cameras on a desktop with Ubuntu 20.04, on 3.1 GHz Intel Xeon w5-3435x CPU and an Nvidia 6000 Ada GPU, and achieves 240 FPS with the same setup on a laptop with Ubuntu 20.04, on 3.2 GHz AMD R7-5800H CPU and an Nvidia GeForce RTX 3060 GPU.

\begin{figure*}[htbp]
    \centering
    \newcommand{\colw}{0.19}
    \newcommand{\figw}{1.0} 
    \includegraphics[width=\figw\textwidth,trim={0cm 0cm 0cm 0cm},clip]{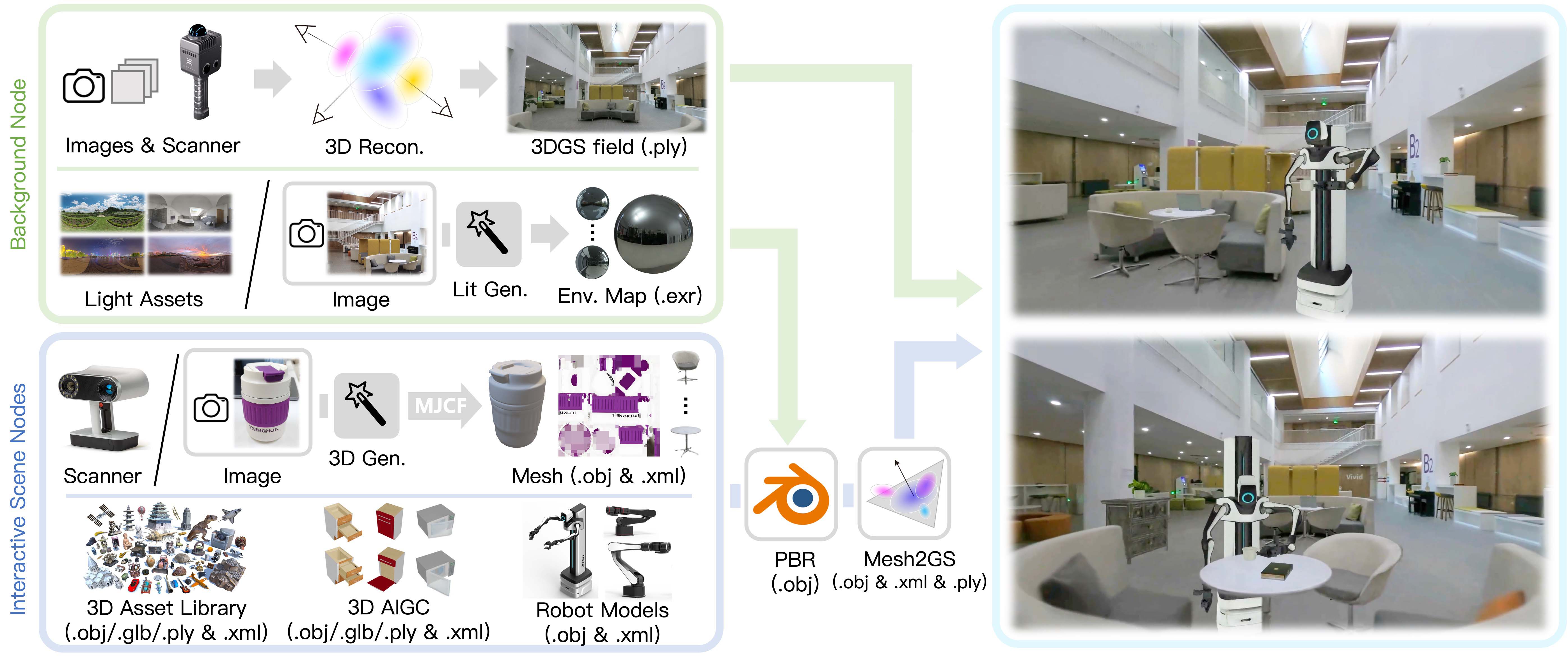}
    \hfill
\vspace{-0.6cm}
\caption{\textsc{Discoverse} Real2Sim generation pipeline. We use 3DGS as a universal visual representation and integrate laser scanning, state-of-the-art generative models, and physically-based relighting to boost the geometry and appearance fidelity of the reconstructed radiance fields. 
}
\vspace{-0.3cm}
\label{fig:real2sim}
\end{figure*}
\section{Real2Sim Pipeline}
\label{sec:real2sim}
We now present the Real2Sim pipeline of \textsc{Discoverse} for high-fidelity asset generation. Our pipeline leverages 3DGS~\cite{kerbl20233d} as a universal visual representation and integrates laser scanning, state-of-the-art generative models, and PBR techniques to boost the photorealism of the reconstructed radiance fields. 

In this section, we first describe the scene-level Real2Sim approach for generating the background node (Sec.~\ref{method:scene}), followed by an introduction to our object-level Real2Sim approach for generating interactive scene nodes (Sec.~\ref{method:object}). 






\subsection{Background Node Real2Sim Generation}
\label{method:scene}
To generate high-fidelity, scene-level assets for use as the background node in simulation, we propose to reconstruct real-world scenes as 3DGS fields with geometry regularization. Additionally, we leverage generative models to recover the environmental lighting, so as to augment the appearance of the interactive nodes for photorealistic rendering. 

{\bf{Real-world Scene Reconstruction.}} Our goal is to robustly digitize real-world complex scenes across various types and scales. However, optimizing 3DGS fields purely with multi-view images oftentimes leads to suboptimal results, e.g., with excessive blurries and floaters due to the degraded geometry. Therefore, we use LixelKity K1 scanner\footnote[1]{https://xgrids.com/lixelk1} to obtain reliable geometry measurements to regularize the radiance fields, similar to~\cite{cui2024letsgo, rematas2022urban, yan2024street}. 

{\bf{Lighting Estimation.}}
\label{method:scene:lit}
Ideally, the interactive scene nodes should be reconstructed independently of the scene and can be seamlessly integrated into any digitized environment (detailed later in Section~\ref{method:object}). To make the appearance of these nodes better align with the reconstructed scene radiance, we propose to estimate the environmental illumination to simulate realistic distant lighting effects with PBR. 

To achieve this, we employ DiffusionLight~\cite{phongthawee2024diffusionlight} to generate the HDR environment map from a single image of the scene, which maintains photorealism while enables plausible variations for randomization.

\subsection{Interactive Scene Nodes Real2Sim Generation}
\label{method:object}
As described in Sec.~\ref{sec:system}, interactive scene nodes require a dual 3DGS-Mesh representation to support both high-fidelity rendering and accurate physics simulation. To make the best use of off-the-shelf tooling, we propose to first work on the classical textured mesh representation for 3D reconstruction, 3D generation, geometry processing, and Pre-PBR. We then develop an efficient Mesh2GS transfer approach to seamlessly integrate the processed mesh into the splatting renderer.

{\bf{Laser-scanned 3D Reconstruction.}}
For objects with approximately Lambertian surface, we use Artec Leo\footnote{https://www.artec3d.com/portable-3d-scanners/artec-leo-fb} scanner to reconstruct the precise 3D geometry and rich textures. Specifically, we place the target object on a turntable and perform multiple scans by rotating and flipping. We then register~\cite{besl1992method} these scans to obtain the complete textured mesh.

{\bf{3D Generation.}}
Real-world objects commonly exhibit non-Lambertian specularity or thin structures, posing great challenges for scanners and traditional image-based 3D reconstruction techniques. Fortunately, recent advances in native 3D generation~\cite{xiang2024structured, zhang2024clay} have demonstrated great success in learning useful geometric priors from high-quality 3D data. We thus choose a recent 3D generation model, CLAY\footnote{https://hyperhuman.deemos.com/rodin}~\cite{zhang2024clay}, to faithfully reconstruct objects with challenging materials or shapes, using a single real-world image as conditional signal. 


{\bf{Relighting.}}
Since it is impractical to capture each object or robot in the scene, the appearance of interactive nodes generally differ a lot from the reconstructed background node due to inconsistent environmental illuminations, resulting in a noticeable gap towards photorealism. To mitigate this issue, we use the estimated HDR environment map (Section~\ref{method:scene:lit}) and Blender\footnote{https://www.blender.org/} to mimic distant lighting effects. 
To prioritize the rendering throughput of \textsc{Discoverse}, we do not perform the conventional PBR online. Instead, we treat PBR as a preprocessing step to align the appearance of the interactive nodes with the background radiance. 

{\bf{Mesh-Gaussian Transfer.}}
So far, we have obtained high-quality geometry and rendering of interactive scene nodes in the form of textured meshes. However, these are not directly compatible with the 3DGS-based background node. To address this, we propose Mesh-Gaussian transfer, to enable a robust and efficient transition between these two representations.

To convert from mesh to 3DGS, we initialize one Gaussian for each mesh facet and locate the centroid of the Gaussian on the barycenter of the respective triangle. We also flatten 3D Gaussian ellipsoids to 2D slanted planar primitives and align the local frame of each Gaussian with the corresponding face normal. 
We initialize the size of the Gaussian by setting the scaling values of the tagent axes to be the mean barycenter-to-vertex distance. During optimization, we incorporate additional geometric constraints in terms of depth and opacity to regularize the 3DGS field.

In cases when conversion from 3DGS to mesh is required, we first render multi-view depth maps from the 3DGS field and then apply TSDF fusion~\cite{newcombe2011kinectfusion} and decimation~\cite{garland1997surface}.

\subsection{Domain Randomization}
\label{method:rand}
To further mitigate domain shifts and account for unmodelled real-world dynamics, we exploit several randomization mechanisms for data augmentation. These include random video overlays (where Internet video crops are randomly selected and linearly blended into the simulation stream), HSV-space image augmentation, and random gamma correction. We also extend the latest generative randomization approach~\cite{yu2024learning}, leveraging ControlNet~\cite{zhang2023adding} for effective conditioning, GPT-4V~\cite{2023GPT4VisionSC} for text-prompt augmentation, and a hybrid flow-based pipeline~\cite{teed2020raft, farneback2003two} for frame interpolation.
\section{Experiments and Applications}
\label{sec:exp}
\begin{table*}[htbp]
\centering
\setlength{\tabcolsep}{0.7mm}{
\caption{Zero-shot {Sim2Real} success rates of ACT~\cite{zhao2023learning} trained on \textsc{Discoverse} and other simulators.}
\label{tab:insitu_sim2real}
\begin{tabular}{c|c|cccc|cccc}
\specialrule{.1em}{.05em}{.05em}
Tasks & Real2Real & MuJoCo~\cite{todorov2012mujoco} & RoboTwin~\cite{mu2024robotwin} & SplatSim~\cite{Qureshi2024SplatSimZS} & \textbf{\textsc{Discoverse}} & $^{\dagger}$MuJoCo~\cite{todorov2012mujoco} & $^{\dagger}$RoboTwin~\cite{mu2024robotwin} & 
$^{\dagger}$SplatSim~\cite{Qureshi2024SplatSimZS} & $^{\dagger}$\textbf{\textsc{Discoverse}} \\ 
\hline\hline
\textit{Close-Laptop} & $100\%$ & $2\%$ & $0\%$ & $56\%$ & $\textbf{66\%}$ & $6\%$ & $0\%$ & $72\%$ & $\textbf{86\%}$ \\
\textit{Push-Mouse} & $94\%$ & $48\%$ & $24\%$ & ${68\%}$ & $\textbf{74\%}$ & $64\%$ & $42\%$ & $74\%$ & $\textbf{90\%}$ \\
\textit{Pick-Up-Kiwifruit} & $100\%$ & $8\%$ & $0\%$ & $26\%$ & $\textbf{48\%}$ & $36\%$ & $0\%$ & $44\%$ & $\textbf{76\%}$ \\
\specialrule{.1em}{.05em}{.05em}
\textbf{Average} & $\textbf{98.5\%}$ & $14.5\%$ & $6\%$ & $44\%$ & $\textbf{55\%}$ & $26.5\%$ & $10.5\%$ & $68\%$ & $\textbf{86.5\%}$ \\
\specialrule{.1em}{.05em}{.05em}
\end{tabular}
}

\begin{minipage}{20cm}
\scriptsize 
${\dagger}$ with image-based data augmentation. We perform random video overlays, HSV-space randomization, and random gamma corrections.
\end{minipage}

\end{table*}
\begin{table*}[htbp]
\centering
\setlength{\tabcolsep}{0.7mm}{
\caption{Zero-shot {Sim2Real} success rates of Diffusion Policy~\cite{chi2023diffusion} trained on \textsc{Discoverse} and other simulators.}
\label{tab:dp_sim2real}
\begin{tabular}{c|c|cccc|cccc}
\specialrule{.1em}{.05em}{.05em}
Tasks & Real2Real & MuJoCo~\cite{todorov2012mujoco} & RoboTwin~\cite{mu2024robotwin} & SplatSim~\cite{Qureshi2024SplatSimZS} & \textbf{\textsc{Discoverse}} & $^{\dagger}$MuJoCo~\cite{todorov2012mujoco} & $^{\dagger}$RoboTwin~\cite{mu2024robotwin} & 
$^{\dagger}$SplatSim~\cite{Qureshi2024SplatSimZS} & $^{\dagger}$\textbf{\textsc{Discoverse}} \\ 
\hline\hline
\textit{Close-Laptop} & $100\%$ & $0\%$ & $0\%$ & $70\%$ & $\textbf{86\%}$ & $0\%$ & $0\%$ & ${90\%}$ & $\textbf{96\%}$ \\
\textit{Push-Mouse} & $96\%$ & $0\%$ & $22\%$ & ${54\%}$ & $\textbf{60\%}$ & $26\%$ & $36\%$ & $82\%$ & $\textbf{88\%}$ \\
\textit{Pick-Up-Kiwifruit} & $94\%$ & $0\%$ & $0\%$ & $12\%$ & $\textbf{22\%}$ & $6\%$ & $0\%$ & $52\%$ & $\textbf{74\%}$ \\
\specialrule{.1em}{.05em}{.05em}
\textbf{Average} & $96.6\%$ & $0\%$ & $7.3\%$ & $45.3\%$ & $\textbf{56\%}$ & $10.6\%$ & $12\%$ & $74.6\%$ & $\textbf{86\%}$ \\
\specialrule{.1em}{.05em}{.05em}
\end{tabular}
}

\begin{minipage}{20cm}
\vspace{0.1cm}
\scriptsize 
${\dagger}$ with image-based data augmentation. We perform random video overlays, HSV-space randomization, and random gamma corrections.
\end{minipage}

\end{table*}

To evaluate the efficacy of \textsc{Discoverse} in bridging the Sim2Real gap, we conduct extensive experiments on imitation learning across three real-world manipulation tasks. We compare the zero-shot Sim2Real success rates of ACT~\cite{zhao2023learning} and Diffusion Policy (DP)~\cite{chi2023diffusion} trained in \textsc{Discoverse}, against those trained in other simulators, including MuJoCo~\cite{todorov2012mujoco}, RoboTwin~\cite{xiang2020sapien}, and SplatSim~\cite{Qureshi2024SplatSimZS}, as well as Real2Real transfer. 
We also report the effects of image-based randomization and outline two exemplar use cases of \textsc{Discoverse}, including navigation and multi-agent coordination, to showcase the versatility of our workflows.

\subsection{Benchmark on Imitation Learning Based Manipulation}
\label{exp:il}

{\bf{Data Collection.}}
Real-world demonstrations are manually collected by a human expert, whereas \textsc{Discoverse} automates this process with motion planners and a gamepad-based state generation approach, which greatly facilitates demonstration by recording the keypoints that describe the relative pose between the robot and object.
To collect 100 demonstrations, it takes \textbf{146} minutes for real-world collection, but only \textbf{1.5} minutes in \textsc{Discoverse}, leading to an \textbf{$\sim$100$\times$} increase in efficiency and demonstrating the scalability of \textsc{Discoverse}.

{\bf{Evaluation Protocol.}} 
We benchmark Sim2Real policy transfer across three contact-rich real-world tasks: 


1) \textit{Close-Laptop}: manipulate the lid and close a laptop;

2) \textit{Push-Mouse}: push the mouse onto the mouse pad; 

3) \textit{Pick-Up-Kiwifruit}: grasp and pick up a kiwifruit;

We use the publicly available ACT~\cite{zhao2023learning} and Diffusion Policy (DP)~\cite{chi2023diffusion} for imitation learning, and we compare against three state-of-the-art simulators: MuJoCo~\cite{todorov2012mujoco}, RoboTwin~\cite{mu2024robotwin}, and SplatSim~\cite{Qureshi2024SplatSimZS}, based on their open-source implementations. For all simulators, we generate {100} demonstrations each task for ACT and 2,000 demonstrations for DP, with random initial gripper states, end-effector positions and orientations. We run 50 trials for each task during testing and use the task success rate ($\%$) as the metric to evaluate Sim2Real and Real2Real performance. All experiments are conducted using an AIRBOT Play robotic arm equipped with an AIRBOT-Gripper-2 and 2 LRCP V1080P cameras (for ego and third-person view), with deployment on an NVIDIA RTX 4090 GPU.

\begin{figure*}[htbp]
    \centering
    \vspace{-0.7cm}
    \newcommand{\colw}{0.19}
    \newcommand{\figw}{1} 
    \vspace{0.7cm}\includegraphics[width=\figw\textwidth,trim={0cm 0cm 0cm 0cm},clip]{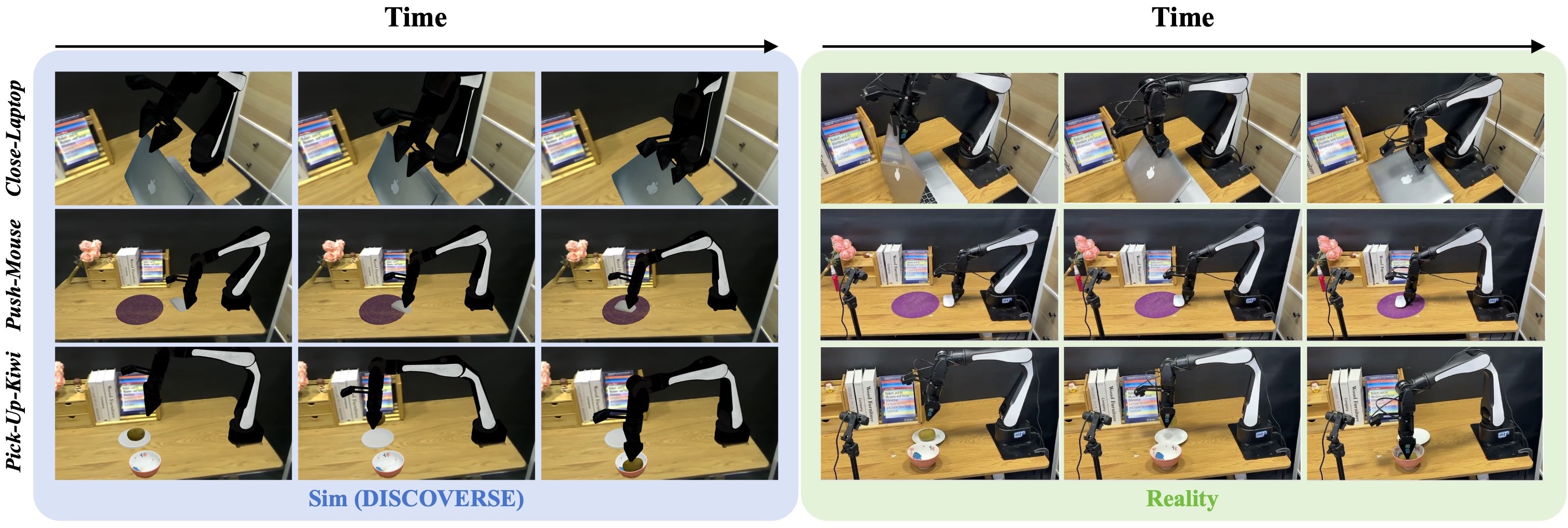}
    \hfill
\vspace{-0.7cm}
\caption{Visualizations of an AIRBOT Play robotic arm performing three different manipulation tasks in the simulation of \textsc{Discoverse} and in reality.}
\label{fig:sim2real}
\end{figure*}
\begin{figure}[htbp]
    \centering
    \newcommand{\colw}{0.19}
    \newcommand{\figw}{0.48} 
    \includegraphics[width=\figw\textwidth,trim={0cm 0cm 0cm 0cm},clip]{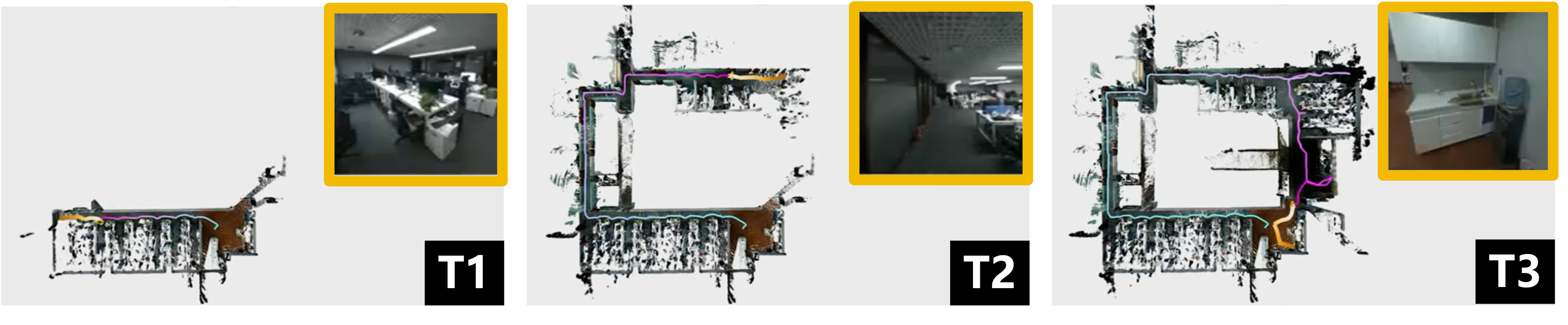}
    \hfill
\vspace{-0.5cm}
\caption{Visualizations of an agent exploring a large-scale indoor scene in \textsc{Discoverse} at different timestamps. The yellow boxes indicate the ego-view inputs generated by the \textsc{Discoverse} renderer. 
}
\label{fig:navigation}
\end{figure}
\begin{figure}[htbp]
    \centering
    \newcommand{\colw}{0.19}
    \newcommand{\figw}{0.48} 
    \includegraphics[width=\figw\textwidth,trim={0cm 0cm 0cm 0cm},clip]{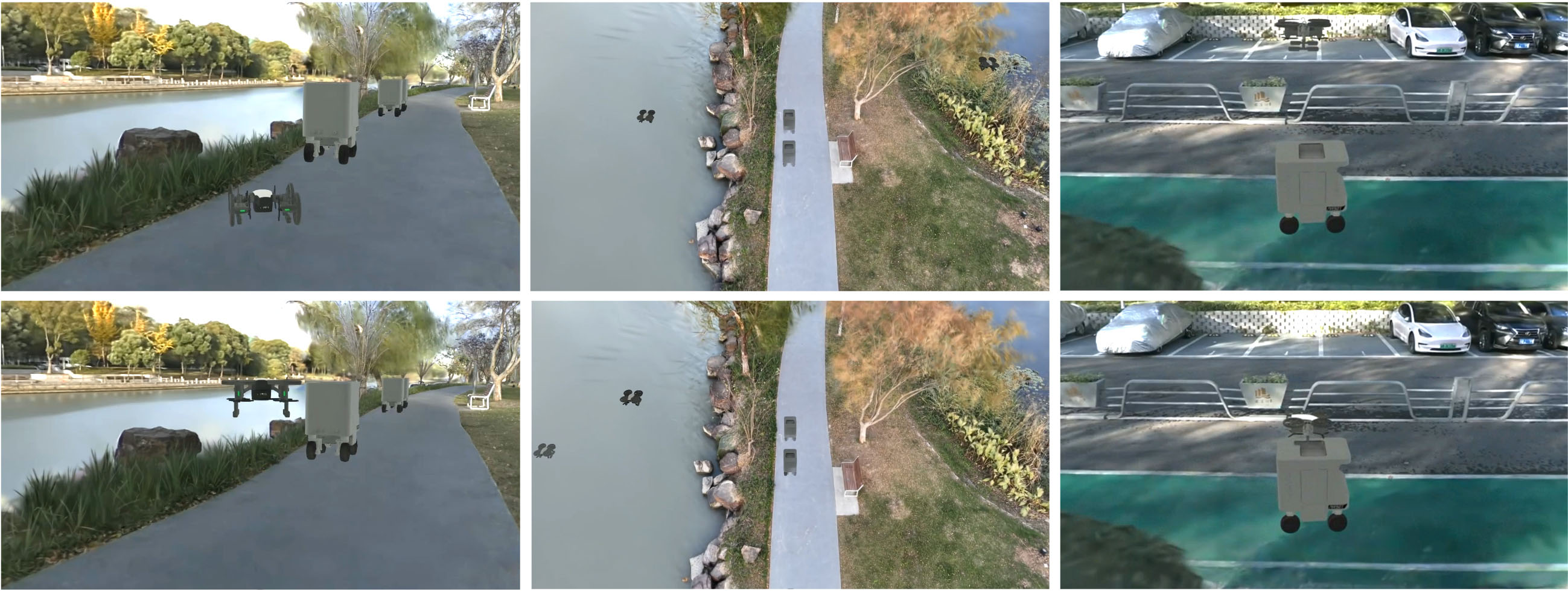}
    \hfill
\vspace{-0.5cm}
\caption{Visualizations of a quadrocopter and a wheeled loco-manipulator cooperatively exploring a large-scale outdoor scene in \textsc{Discoverse}.
}
\label{fig:coordination}
\end{figure}

{\bf{Sim2Real Benchmark.}}
The comparisons on Sim2Real success rates between \textsc{Discoverse} and other simulators are presented in Table~\ref{tab:insitu_sim2real} (ACT) and Table~\ref{tab:dp_sim2real} (DP). Policies trained on MuJoCo~\cite{todorov2012mujoco} and RoboTwin~\cite{mu2024robotwin} struggle with zero-shot Sim2Real transfer for both ACT and DP, primarily due to the significant domain shifts resulting from the poor visual fidelity. While SplatSim~\cite{Qureshi2024SplatSimZS} partially addresses this issue with 3DGS-based Real2Sim, \textsc{Discoverse} achieves the best Sim2Real results due to superior fidelity, outperforming SplatSim by $\sim$11\% on the average task success rate for both ACT and DP, without any data augmentation.

The Sim2Real results with image-based data augmentation are presented in the right sections of Table~\ref{tab:insitu_sim2real} and Table~\ref{tab:dp_sim2real}. All simulators show notable improvements after randomization, and \textsc{Discoverse} achieves a 31.5\% increase on the average success rate for ACT and a 29.3\% improvement for DP. Remarkably, \textsc{Discoverse} outperforms SplatSim by 18.5\% for ACT and by 11.4\% for DP, when employing the same data augmentation mechanisms.


\subsection{Exemplar Applications}
\label{exp:applications}
\textsc{Discoverse} is a unified and versatile simulator, with extendable supports for a variety of robot models and downstream tasks. We now showcase two exemplar applications in the context of navigation and multi-agent coordination.

{\bf{Navigation.}} We deploy an AIRBOT MMK2 agent to navigate a large-scale indoor scene in \textsc{Discoverse}, following predefined key points. The agent takes ego-view renderings as input and progressively updates the spatial map~\cite{li2024activesplat}. Fig.~\ref{fig:navigation} illustrates the visual inputs and the reconstructed spatial map at different timestamps during the exploration.

{\bf{Multi-Agent Coordination.}} As another example shown in Fig.~\ref{fig:coordination}, we deploy a wheeled loco-manipulator and a quadrocopter for cooperative exploration of a large-scale in-the-wild scene with \textsc{Discoverse}. The quadrocopter can transform to a wheeled mode, enabling it to land on the wheeled loco-manipulator or the ground for efficient delivery in hard-to-reach areas.
\section{Conclusion and Futher Work} 
\label{sec:conclusion}
In this work, we introduce \textsc{Discoverse}, the first unified, modular, open-source 3DGS-based simulation framework to bridge the Sim2Real gap in robot learning. By integrating Gaussian Splatting and MuJoCo, \textsc{Discoverse} enables hyper-realistic simulation of complex real-world scenarios, with inclusive supports for various sensor modalities, existing 3D assets, robot models, ROS plugins, and various randomization approaches. These features make \textsc{Discoverse} ideal for large-scale robot learning, efficient data synthesis, and complex robotic benchmarks, e.g., manipulation, navigation, multi-agent coordination, etc. Through extensive experiments on imitation learning with ACT and DP, we verify the superiority of \textsc{Discoverse} in closing the Sim2Real gap, compared to existing simulators. 

{\bf{Future Work.}} We believe \textsc{Discoverse} represents the beginning of a new era in Real2Sim2Real robot learning, paving the way for end-to-end optimization across the entire Real2Sim and Sim2Real pipeline. While Sim2Real in IL is just the beginning, we envision the capability of \textsc{Discoverse} to facilitate the zero-shot Sim2Real transfer for more complex RL policies beyond IL, thus alleviating the onerous process of real-world RLPF. Future improvements for \textsc{Discoverse} will focus on advanced physical simulation, Mesh-GS hybrid PBR, etc. With \textsc{Discoverse}, we plan to establish a variety of Sim2Real benchmarks for end-to-end robot learning for stimulating further research and practical applications in the field.
\section{Acknowledgement} 
\label{sec:acknowledgement}
The authors gratefully acknowledge the support of D-Robotics under Grant 20243000104. The authors would also like to acknowledge DISCOVER Lab and DISCOVER Robotics for technical and hardware supports; Liang Zhu for optimizing the user interface; Shihui Zhou for integrating imitation learning algorithms.

\bibliographystyle{IEEEtran}
\bibliography{ref}

\end{document}